# SelfFed: Self-Supervised Federated Learning for Data Heterogeneity and Label Scarcity in Medical Images

Sunder Ali Khowaja[1], Kapal Dev[2], Syed Muhammad Anwar[3,4], Marius George Linguraru[3,4]

[1] Department of Telecommunication Engineering, Faculty of Engineering and Technology, University of Sindh, Jamshoro, Sindh, Pakistan

[2] Department of Computer Science and ADAPT Centre, Munster Technological University, Bishopstown, Cork, T12 P928, Ireland,

[3] Sheikh Zayed Institute for Pediatric Surgical Innovation, Childrens National Hospital, Washington D.C., United States of America

[4] Department of Radiology and Pediatrics, George Washington University, School of Medicine and Health Sciences, Washington D.C., United States of America

Sandar.ali@usindh.edu.pk, kapal.dev@mtu.ie, sanwar@childrensnational.org, MLingura@childrensnational.org

**Abstract**

Self-supervised learning in the federated learning paradigm has been gaining a lot of interest both in industry and research due to the collaborative learning capability on unlabeled yet isolated data. However, self-supervised based federated learning strategies suffer from performance degradation due to label scarcity and diverse data distributions, i.e., data heterogeneity. In this paper, we propose the SelfFed framework for medical images to overcome data heterogeneity and label scarcity issues. The first phase of the SelfFed framework helps to overcome the data heterogeneity issue by leveraging the pre-training paradigm that performs augmentative modeling using Swin Transformer-based encoder in a decentralized manner. The label scarcity issue is addressed by fine-tuning paradigm that introduces a contrastive network and a novel aggregation strategy. We perform our experimental analysis on publicly available medical imaging datasets to show that SelfFed performs better when compared to existing baselines and works. Our method achieves a maximum improvement of 8.8% and 4.1% on Retina and COVID-FL datasets on non-IID datasets. Further, our proposed method outperforms existing baselines even when trained on a few (10%) labeled instances.

## 1. Introduction

Internet of Medical Things (IoMT) and medical imaging have been active and challenging areas for the researchers associated with machine learning. Medical imaging and IoMT comprise various tasks including segmentation, classification, and disease detection (Khowaja et al., 2018). Standard machine learning algorithms for analyzing data generated by IoMT have been proven to be successful in improving the performance of aforementioned tasks. However, there are some challenges associated with medical imaging applications that cannot be tackled with traditional machine learning methods. Among these, data privacy is one of the major challenges due to the patient's data being shared to train a machine learning algorithm. The data sharing problem not only risks data leakage but is also vulnerable to data manipulation, resulting in a wrong diagnosis. These vulnerabilities include differential privacy and adversarial attacks (Khowaja et al., 2022). Another challenge corresponds to the learning of IoMT data in a distributed manner. With heterogeneous sensing devices and wearable sensors, the IoMT data is rarely stored on a centralized server, therefore, data from different servers placed at different hospitals need to be trained in a distributed manner. Further, in real scenarios, the data from different servers and hospitals probably follow different distributions (C. He et al., 2020). In this regard, a decentralized learning paradigm is needed to train the data on client devices in order to ensure privacy and tackle heterogeneity in data distribution. The data privacy and varying data distribution issues have been addressed by a seminal work (McMahan et al., 2017), which proposed the use of federated learning (FL) paradigm. In FL, each client would train individual models locally on the client side, followed by an aggregation of client models on the server side to construct a global model. The copies of the global model are then sent back to the client for improved inference.

In recent years, FL has seen drastic growth due to its characteristics that break regulatory restrictions, reduce data migration costs, and strengthen data privacy. The FL technique is characterized as a distributed machine learning technique and has been widely adopted for applications in the domain of data mining, natural language processing, and computer vision. However, FL implementations cannot deal with data heterogeneity, which is one of the challenges that resists the realization of wider adoption of FL methods. Data heterogeneity issue gets more challenging due to the generation of diversified datasets with varying distributions (due to heterogeneous sensing devices) and behavioral preferences at the client side that result in a skewed distribution of labels (some sensors could use heart rate, while others use heart rate variability or blood volume pulse in their daily life). Another challenge associated with the FL approach is label scarcity which has been explored relatively less in the available literature. This issue could get severe in the edge device setting, as the users do not have a user-friendly interface to annotate their data on their IoT or smartphone device. Further, in some cases, users are resistant to annotating sensitive and private data due to privacy concerns. This is one of the probable reasons why large-scale labeled medical image datasets are scarce in availability.

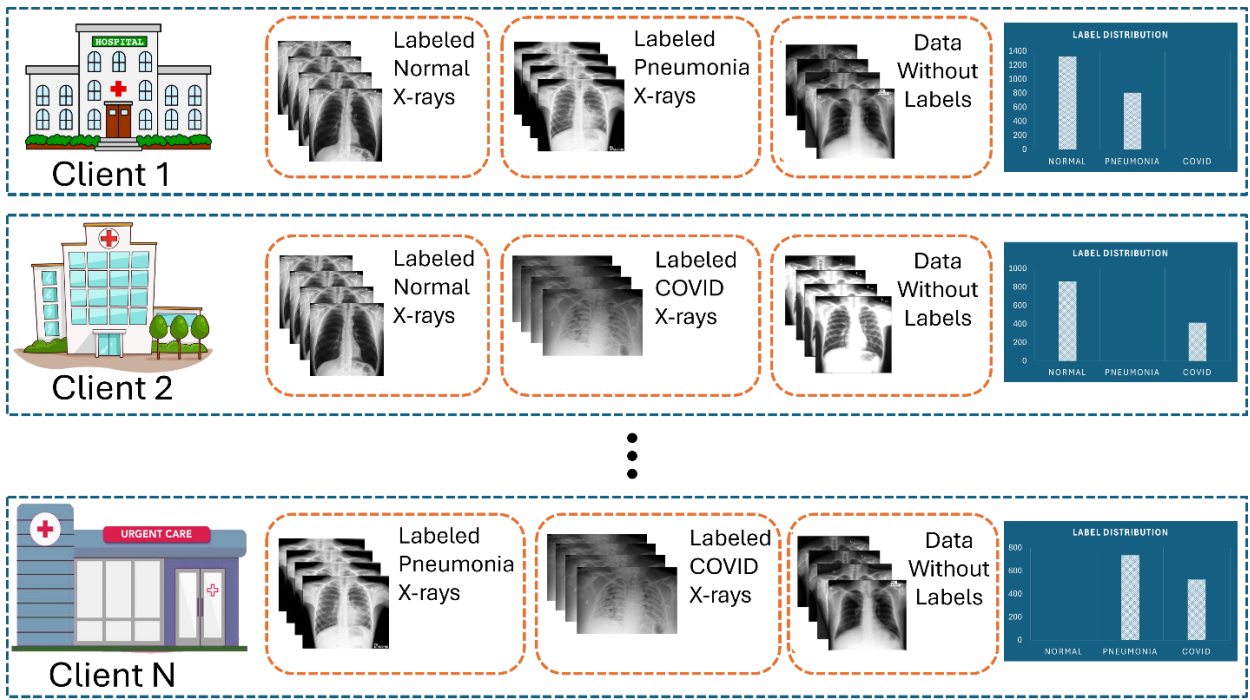

Figure 1 A federated learning scenario highlighting label scarcity and IoMT data heterogeneity issues from various clients.

Researchers have worked extensively on the mitigation of data heterogeneity issues by proposing the training of a global model such as FedOPT, FedNOVA, FedProx, and FedAvg. Researchers have also proposed distributed and personalized FL frameworks to deal with the issue of data heterogeneity in the form of DBFL, Per-FedAvg, Ditto, and pFedMe (Khowaja et al., 2021; Li et al., 2020; McMahan et al., 2017; Reddi et al., 2020; J. Wang et al., 2020). However, the aforementioned methods work under the assumption that sufficient labels are available at the edge, which is not true in real-life scenarios, especially in the case of medical imaging. An Example of label scarcity and data heterogeneity in medical images is shown in Figure 1. For instance, some hospitals have more data concerning the severe conditions of the patients while others have more data related to the normal conditions of the patients. This phenomenon is referred to as label distribution skew or data partitions with non-identical distribution (non-IID). Another problem highlighted in Figure 1 is related to quantity skew as some hospitals have large amounts of data while others have a smaller number of images. Similarly, each of the client hospitals might use different patient populations, acquisition protocols, and devices, which results in feature distribution skew. Label Scarcity is also one of the challenges represented in Figure 1 as some hospitals have large numbers of annotated data while others might lack expertise for annotating the data or a can make a small proportion of labeled data available, respectively.

Recently, researchers proposed to tackle the label scarcity issue in FL through semi-supervised methods by assuming that either client or server possesses a few labeled instances (Itahara et al., 2023). The semi-supervised methods are based on pseudo-labeling and consistency loss for training a global model. Researchers also tried using unsupervised learning using the Siamese network, but the method was not able to handle the data heterogeneity issue quite well. Furthermore, unsupervised learning in FL was also limited in terms of scalability concerning IoMT devices. Researchers have started exploring self-supervised learning strategies such as SimSiam (X. Chen & He, 2021), BYOL (Grill et al., 2020), SwAV (Caron et al., 2020), and SimCLR (T. Chen et al., 2020) to tackle the problem of label scarcity but the problem of data heterogeneity remains at large.

Extensive research efforts have been carried out to solve skewed data distribution problems that are caused due to data heterogeneity. A recent study suggested that the family of vision transformers (Qu et al., 2022) can reduce the issue of data heterogeneity to a certain extent in comparison to traditional convolutional neural networks (CNNs). The study by (Qu et al., 2022) showed that adopting vision transformers can improve the overall performance of FL methods. Additionally, the statement holds on general computer vision applications that rely on supervised pre-training from ImageNet the same cannot be adopted to IoMT data due to domain discrepancy. Therefore, a method that can handle data heterogeneity and label scarcity by leveraging small-scale labeled instances on the client side and large-scale unlabeled data on the server side to learn the global model in a collaborative manner is needed.

In this paper, we address the label scarcity and data heterogeneity issue by proposing self-supervised federated learning (SelfFed) using online learning principles. The framework employs two phases, i.e., the SelfFed pre-training phase and the SelfFed fine-tuning phase. Studies have proved that pre-training pertaining to self-supervised learning is an effective strategy to deal with inductive bias and skewness in label distribution. The visual features learned from unlabeled data on the client side are leveraged to update the online network encoder on the server side. A contrastive network using consistency loss is trained on the server side, which shares the encoder weights and model parameters with the client side for stage 2 proceedings. The client-side model in stage 2 relates to a linear classifier, which is then trained in a supervised manner. The encoder parameters updated via cross-entropy loss are then shared with the online network encoder for further aggregation. A Swin Transformer Encoder has been employed for extracting effective representations to achieve better performance. Unlike the existing works, we use a target network and projection heads to extract high-quality representations. To the best of our knowledge, the framework with online learning principles, Swin Transformer Encoder, and novel aggregation mechanism that could deal with both label scarcity and data heterogeneity for medical image applications has not been proposed. The contributions of this work are summarized as follows:

- We propose a novel framework SelfFed, which simultaneously deals with both data heterogeneity and label scarcity for medical images.
- We further propose a novel contrastive network and aggregation mechanism to perform fine-tuning in a self-supervised FL paradigm.
- We propose the use of a Swin transformer along with MAE to perform augmentative modeling as a self-supervised task.
- We evaluate the proposed method on various datasets and show that SelfFed performs better than existing FL methods along with supervised baselines pre-trained on ImageNet using non-IID data.

## 2. Related Works

This section consolidates the review of articles concerning traditional self-supervised learning techniques that include contrastive learning, followed by the review of techniques based on the FL paradigm. Lastly, we review some works that combine FL and self-supervised learning for a particular application.

2.1 Contrastive Learning

In recent years, self-supervised learning has gained a lot of attention due to its capability of utilizing pretext task information for extracting better features and representations that can be leveraged as supervised information (Hervella et al., 2020). Therefore, it is important to come up with a relevant pretext task that might vary with respect to different domains including graphs, computer vision, natural language processing, and so forth. Generally, self-supervised techniques are categorized into contrastive-based, temporal-based, and context-based methods. The temporal-based methods model the temporal characteristics of the data such as next sentence prediction in natural language processing through BERT, or the next action to be carried out in a video while computing the frame similarity (Sermanet et al., 2018). The sequential property in temporal-based self-supervised learning is also referred to as an order-based constraint. Another technique utilizes contextual information such as the frequency of words or similar words that define the same thing through Word2Vec for the task construction. This technique is termed context-based self-supervised learning. Many researchers have used contextual information such as image coloring, rotation prediction, jigsaw, and other representations, for the tasks in computer vision field. However, contrastive learning is the most popular technique to be used with self-supervised learning.

Contrastive learning takes into account the feature space to distinguish between the data rather than considering the monotonous details of instances. This helps contrastive learning to generate a model that is more generalizable and simpler. Contrastive learning uses InfoNCE loss (Oord et al., 2018) to learn discriminating features among contrasting instances while identifying supportive features for identical instances. The simplicity and generalization of the model can be achieved as contrastive learning is able to distinguish or identify similarity in instances at an abstract semantic level, i.e., feature spaces. One of the ways to improve the performance of contrastive learning is to expand upon the number of negative samples. One of the seminal studies that employ the Siamese network for maintaining a queue of negative samples through a momentum encoder is MoCo (K. He et al., 2020). On the other end of the spectrum, SimCLR (T. Chen et al., 2020) technique undertakes negative samples available in the current batch. Another approach has been proposed, i.e., BYOL (Grill et al., 2020) that does not require negative samples, rather it uses an augmented view of the available images to train its online networks while maintaining a decent performance. Out of the aforementioned three, BYOL represents the closest match to the real-world situation.

Recently, vision transformers (ViTs) have been adopted for federated learning (Qu et al., 2022) such as MAE (K. He et al., 2022), BEiT (Bao et al., 2022), and Swin Transformers (Liu et al., 2021) to overcome the data heterogeneity issues. Furthermore, the ViTs are also capable of learning representations from corrupted images using signal reconstruction. We refer to these methods as augmentative modeling. Where contrastive learning relies on data augmentations or a large sample size, the ViTs achieve the same performance level due to their intrinsic characteristics.

## 2.2 Federated Learning

Federated learning was first introduced by Google (McMahan et al., 2017) and has been popular due to its privacy preservation characteristics. FL paradigm eliminates the requirement of data sharing from the client side while allowing models to train in a collaborative manner. Generally, FL approaches train the model locally on the client side and performs the aggregation of a global model on the server side. The aggregation and optimization of the global model are carried out in an iterative manner through communication rounds between the server and the client, accordingly. The flow of the operations in FL is as follows:

- A global model from the server is sent to the clients for collaborative training.
- Clients update the global model locally with their local training data and send the updated model to the server.
- Model updates from clients are aggregated at the server for the update. Once the model is updated, it can be used for various tasks.

There have been many algorithms in the FL paradigm to aggregate the model updates, however the most popular of them is the federated averaging algorithm (FedAvg) (McMahan et al., 2017). Local parameters are updated through multiple gradient descent updates, which increases the computation at the local node but decreases the amount of communication. Another similar approach for aggregating model updates is FedSGD (McMahan et al., 2017). The main problem with such methods is its convergence that requires a large number of iterations, along with the homogeneity of data and devices involved. Various methods have been proposed over the years to address non-IID data issues. These approaches can be generally categorized as methods for improving aggregation and methods focusing on stability. The former strives for improving the efficiency of model aggregation processes such as FedNova (McMahan et al., 2017), while the latter pivots for the stability of the training process at the local nodes. The methods concerning stabilization include FedAMP (Huang et al., 2021), and FedProx (Li et al., 2020).

The problem with the existing FL approaches is the assumption that the true labels for all the instances are available on devices. However, labeled data is a limited resource and requires expertise when performing labeling. Even in some domains such as medical imaging, labeling is synonymous with high costs. Another problem is associated with the heterogeneity of data from a model initialization perspective (H.-Y. Chen et al., 2022). Some studies suggest that the problem of data heterogeneity can be alleviated by performing pre-training, however, the use of ViTs has been considered to be a more viable solution to address data heterogeneity issues (Qu et al., 2022). Nevertheless, it is imperative to design FL-based methods that deal with both issues, i.e., data heterogeneity and label scarcity, respectively.

## 2.3 Federated Learning Based on Contrastive Networks

Contrastive networks employed in the FL paradigm have shown promising results in alleviating non-IID problems (P. Wang et al., 2021). The use of contrastive networks in FL is generally performed in two phases. The first phase extracts visual representations from distributed devices by performing collaborative pre-training on unlabeled data. This phase forwards the shared features and performs aggregation on model updates similar to conventional FL approaches. On the client side, local unlabeled data is utilized to perform comparative learning. The second phase performs fine-tuning using a limited number of labels and a pre-trained model from the previous phase. The second phase process is carried out via federated supervised learning on each device, respectively.

The study in (Zou et al., 2024) proposed a SSFedMRI that addresses data deficiency issues and data heterogeneity using collaborative training and physics-based reconstruction method. However, most focus of the study is on the reconstruction of MRI images and reduction of communication costs, which is quite different from the work that is being proposed in this study. The study in (Zhang et al., 2024) proposed a Barlow Twins based FL algorithm that is based on contrastive learning strategy to learn common inherent representations and center-specific information for the optimization of FL training. The study heavily leverages generative adversarial networks (GANs) for synthetic data generation to train the pretext task. The method considers histopathological images for the the validation and evaluation of their proposed approach. The study does not put its focus on heterogeneity or data scarcity issues. The study in (Elbatel et al., 2023) proposed a dynamic balanced model aggregation method for classification of anomalies from gastrointestinal and skin lesion images. The aggregation was performed using self-supervised priors. Although the study focuses on label skewness, the main contribution of this study was the design of global aggregation method. Wang et al. (R. Wang et al., 2024) in their study focused on the data augmentation strategy by adding noise to the labeled samples for overpopulating the data. The study was limited to tabular datasets only. Furthermore, the study did not undertake the issues of label scarcity and data heterogeneity. The study in (Yang et al., 2024) proposed a dense contrastive-based federated learning method for medical image prediction tasks. The focus of this study was to design a contrastive network that could optimize the representation agreement through multiscale representations during the aggregation process. The study focused on data heterogeneity issues while considering the pulmonary nodule detection dataset. The study in (Zheng et al., 2024) proposed the usage of variational autoencoders (VAEs) for knowledge distillation, data generation, and the extraction of latent representation within contrastive learning setting. The study only focused on label scarcity issues, which was handled by the VAEs. The study in (Guan et al., 2024) conducting a thorough survey on the studies that use FL approaches for medical image analysis. The study very briefly highlighted some of the methods that used contrastive learning setting and the methods that deal with label scarcity. However, these studies were highlighted and discussed in a disjoint manner. The study in (Yan et al., 2023) proposed a contrastive learning FL method for the medical image classification task. It is one of the closest to our proposed study, i.e. it deals with both the label scarcity and data heterogeneity issues. However, unlike the proposed method, the study uses ViTs as encoder and does not employ online learning characteristics for their model aggregation and fine-tuning strategy. Technically there are other differences as well, but the aforementioned two distinctions are the main highlights. The comparative table for distinguishing the aforementioned studies from the proposed one is shown in Table 1. The comparison is performed based on the following characteristics: Label Scarcity, Data Heterogeneity, Augmentative tasks, Encoder, and Online Learning, respectively.

In addition, the works employing contrastive networks in the FL paradigm undertake skewness concerning label distributions only. For instance, the number of images is the same with each client, and the number of data instances is usually considered to be more than 10,000. Such an assumption is not always true in real-world settings; therefore, it is yet to be observed how such methods perform when either of the aforementioned assumptions does not hold. In this work, we propose the use of a Swin transformer along with MAE (K. He et al., 2022) to perform augmentative modeling as a self-supervised task. We assume that the use of the Swin transformer and MAE will help the contrastive network in FL converge faster with a smaller number of data instances.

Table 1 Characteristic Comparison with the Existing works to highlight specific novel aspects of the proposed method.

| Study | Year | LS | DH | AT | Enc | OL |
|---|---|---|---|---|---|---|
| Zou et al. | 2024 | ✘ | ✔ | ✘ | CNN Denoiser | ✘ |
| Zhang et al. | 2024 | ✘ | ✘ | ✔ | Generative Adversarial Networks | ✘ |
| Elbatel et al. | 2023 | ✘ | ✔ | ✔ | MoCo-V2 | ✘ |
| Wang et al. | 2024 | ✘ | ✘ | ✔ | CNN | ✘ |
| Yang et al. | 2024 | ✘ | ✔ | ✔ | CNN | ✘ |
| Zheng et al. | 2024 | ✔ | ✘ | ✔ | CNN | ✘ |
| Guan et al. | 2024 | ✘ | ✘ | ✘ |  | ✘ |
| Yan et al. | 2023 | ✔ | ✔ | ✔ | Vision Transformers | ✘ |
| Proposed Method | 2024 | ✔ | ✔ | ✔ | Swin Transformer | ✔ |

## 3. Methodology

3.1 Problem Definition

The main motivation of this work is to design a self-supervised federated learning (SelfFed) approach that does not require data sharing while learning in a decentralized manner. SelfFed is designed in a way such that it should perform well even on non-IID data from clients with a limited number of available labels. The number of clients in this work is denoted by $M$. A local dataset $\mathcal{D}^m$ is associated with each client such that $m \in 1, \ldots, M$. A global model needs to be learned in a generalized manner from all local datasets such that $\mathcal{D} = \bigcup_{m=1}^{M} \mathcal{D}^m$. The empirical loss concerning clients over data distribution $\mathcal{D}^m$ is defined in equation 1.

$$loss_m(w) = \mathbb{E}_{x \sim \mathcal{D}^m}[\mathbb{L}_m(w; x)] \tag{1}$$

The above equation $w$ represent the parameters to be learned concerning the global model, while the notation $\mathbb{L}_m$ refers to the loss function concerning each client $m$. The aforementioned loss is then used to optimize the global objective function as $argmin_w loss(w) = \sum_{m=1}^{M} \frac{|\mathcal{D}^m|}{|\mathcal{D}|} loss_m(w)$. This work focuses on dealing with data heterogeneity problems such that the data can be non-IID, i.e., $\mathcal{D}^a \neq \mathcal{D}^b$ and might follow non-identical data distribution, $PD_a(x, y)$ and $PD_b(x, y)$. This work also undertakes the problem of label scarcity at the local client level. Therefore, we define the unlabeled and labeled dataset as $\mathcal{D}_{ul}^m = x$ and $\mathcal{D}_{lb}^m = (x, y)$, respectively. We also assume that $|\mathcal{D}_{lb}^m|$ is comparatively very small.

3.2 SelfFed Framework

The workflow of the proposed SelfFed framework is shown in Figure 2. In the subsequent sections, we show that the proposed SelfFed framework is able to handle data heterogeneity as well as label scarcity problem. As shown in Figure 2, the proposed framework is divided into two stages: the pre-training and fine-tuning stages concerning self-supervised federated learning. In the pre-training stage, augmentative modeling is exploited in order to learn representative knowledge in a distributed manner. In the latter stage, federated models are fine-tuned by transference of representative knowledge from the previous stage to perform target tasks, respectively. For augmentative modeling, we leverage MAE (K. He et al., 2022) method into our SelfFed framework. The reason for choosing MAE over BEiT is performance advantage as highlighted in (Yan et al., 2023). The MAE-integrated SelfFed framework is denoted by SelfFed-MAE, accordingly. We provide details for the pre-training and fine-tuning of SelfFed-MAE in the subsequent subsections.

## 3.3 SelfFed Pre-Training

Each client consists of a local autoencoder, and a decoder represented as $Enc_m$ and $Dec_m$, during the pre-training phase, respectively. Augmentative modeling is used to train the models, which undertake patches from a subset of an image and learn to reconstruct the marked patches, accordingly. MAE (K. He et al., 2022) is leveraged as our self-supervision module, specifically for performing augmentative modeling. For our IoMT use case, we consider medical images to validate the proposed SelfFed framework.

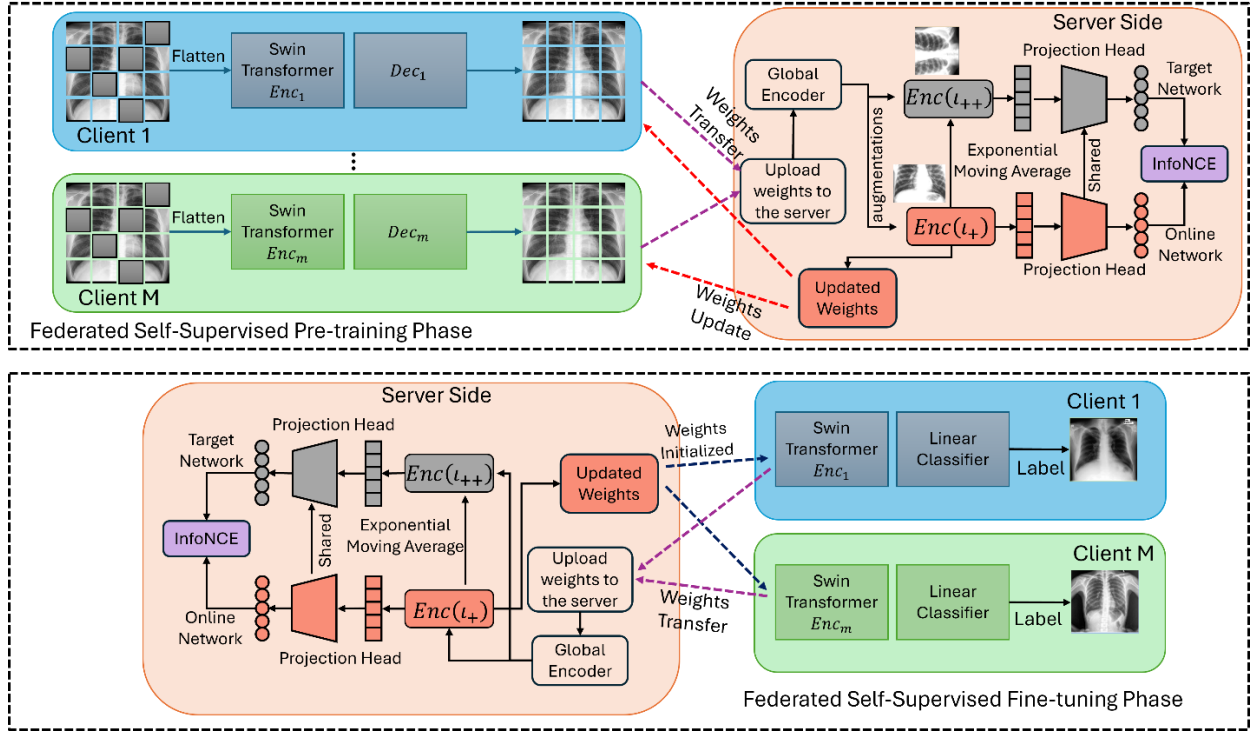

Figure 2 Generalized SelfFed Framework. Augmentative modeling is performed in the first phase, which is referred to as self-supervised pre-training. The process of pre-training comprises three steps. At each communication round, a local auto-encoder and decoder are trained on the client side on unlabeled data. The weights of the local autoencoder are shared with the encoder on the server side. On the server side, a target and an online network is deployed having an encoder and projection head. The online network is responsible for updating the parameters of the target network via an exponential moving average. The server shares the online network's encoder weights with the clients. In the second phase, the client encoders are initialized with the global encoder weights, and a linear classifier is stacked, accordingly. The fine-tuning is performed on labeled data as the client sends only the encoder part to the online network for aggregation.

An input image $x \in \mathbb{R}^{H \times W \times C}$ belonging to a certain data distribution $\mathcal{D}^m$ is segmented into image patches such that $x_r = \{x_r^j\}_{j=1}^{R} \in \mathbb{R}^{R \times (V^2 \cdot C)}$, where $C$ refers to the number of channels. The dimensions of the original and image patch are represented by $(H, W)$ and $(V, V)$, respectively. The number of patches is determined using $R = \frac{HW}{V^2}$. For instance, if the original image has the dimension of 256 x 256, and the image patch has the dimension of 64 x 64, therefore, the number of patches in which the original image will be divided would be 16.

### 3.3.1 Augmentative Masks

The masking ratio for generating augmentation is denoted as $\psi$, the unmasked positions are denoted as $\varsigma$ while the masked positions are denoted as $\vartheta$. When $\psi\%$ of patches are masked in a random manner, we get $\|\varsigma\| + \|\vartheta\| = R$ and $\|\vartheta\| = \psi R$. The representation of overall image patches can be given as shown in Equation 2.

$$x_r = \{x_r^j, j \in \varsigma\} \cup \{x_r^j, j \in \vartheta\} \qquad (2)$$

where the first and second terms in the aforementioned equation characterize unmasked and masked visible patches, respectively. The adopted method MAE (K. He et al., 2022) considers random masking. The process of generating augmentative masks is shown in Figure 3.

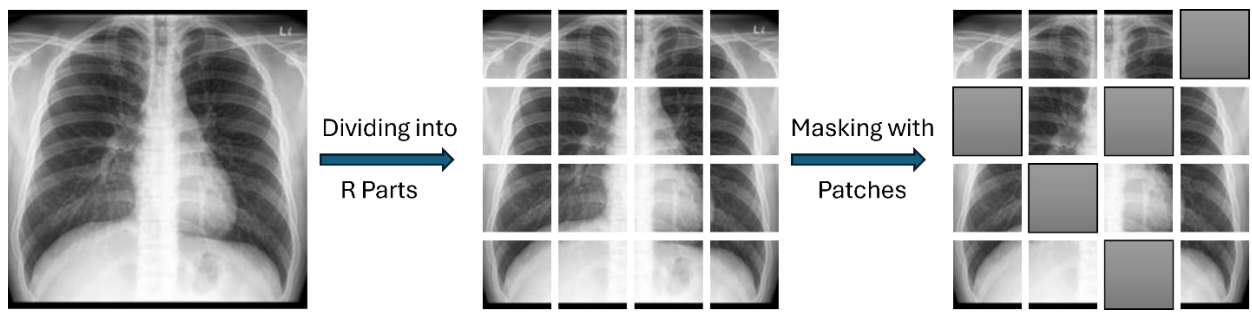

Figure 3 Transforming an image to augmentative masks.

### 3.3.2 Encoder

We employ Swin Transformer (Liu et al., 2021) as the representational encoder in the SelfFed framework, which undertakes the image patch sequences as illustrated in Figure 3. The masked patches are considered as tokens for the Encoder; thus, the Swin Transformer learns to replace the masked tokens via a shared learnable vector rather than removing the mask tokens from images. For the Swin Transformer encoder, we utilized relative and absolute position encoding to get the performance gain. In conjunction with the Swin Transformer encoder, the MAE considers the linear projections with the position embeddings as an input extracted from the patches. The projection is denoted as $x_r^\varsigma Enc$. The output is obtained in the form of an encoded visible patch, which is represented as $\{o_j, j \in \varsigma\}$.

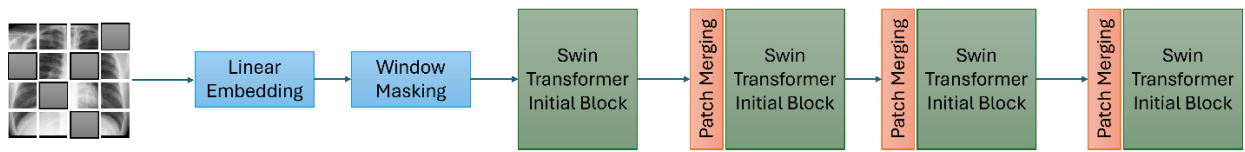

Figure 4 The encoder design using Swin Transformer-based MAE. The window masking retains the masked tokens while bypassing position embedding. Considering the image dimensions of 256 x 256 x 3, the patch partitions are constructed with 64 x 64 x 48. The linear embedding and window masking will yield the resultant dimension of 64 x 64 x 96. The first Swin transformer block will also yield the same dimension, followed by 32 x 32 x 192, 16 x 16 x 384, and 8 x 8 x 768, for each of the subsequent Swin transformer blocks, respectively. This is an exemplary illustration to explain the process.

### 3.3.3 Decoder

Based on the input patches and their encoded representations, the decoder's task is to reconstruct the signal, effectively. The decoder also uses Swin Transformer as the backbone. The architecture of the decoder is quite similar to that of Swin-UNet decoder (Cao et al., 2023). To make the output of encoder compliant with the input of the decoder, we neither remove the class tokens nor add the masked tokens in the decoder. In contrast to the Swin UNet, we employ a projection prediction layer for the image reconstruction part. Furthermore, we do not add any skip connection layer between the decoder and the encoder. The architecture of the decoder is shown in Figure 5. The visible patches are provided as an input to the MAE in addition to the linear embedding and masked patches, i.e., $\gamma_r^\vartheta = \{\gamma_r^j, j \in \vartheta\}$, which acts as a learnable vector for the decoder.

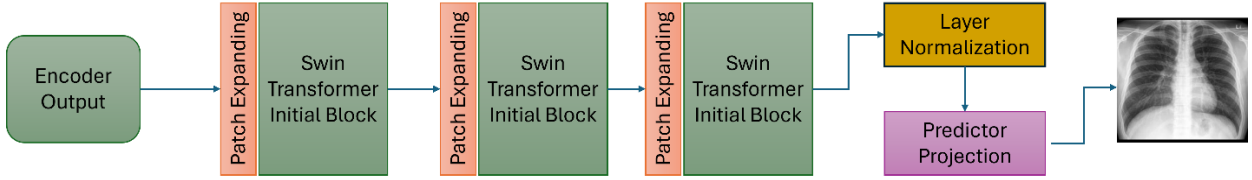

Figure 5 The decoder design using Swin Transformer-based MAE. The encoder output was 8 x 8 x 768 from the encoder output. Each of the Swin Transformer blocks yields the resultant dimensions of 16 x 16 x 384, 32 x 32 x 192, and 64 x 64 x 96, respectively. The layer normalization yields 64 x 64 x 48. The output will be the predicted masks of the image size 256 x 256 x 3.

### 3.3.4 Loss Function

The local objective function through which $Enc_m$ and $Dec_m$ are trained on a local dataset $\mathcal{D}^m$ is given in equation 1. Concerning MAE, the mean square error for the masked patches such that $\{x_r^j, j \in \vartheta\} \in \mathbb{R}^{V \times V \times C}$ is represented by $\mathbb{L}_m$ and can be defined as shown in equation 3.

$$\mathbb{L}_m = \Sigma_{j \in \vartheta} \frac{1}{\|\vartheta\|} \left( \left( x_r^j - \hat{x}_r^j \right)^2 ; w \right) \quad (3)$$

### 3.4 SelfFed Fine-Tuning

On the server side, we have two networks, namely online and target networks that consist of an encoder and a projection head. The input image to the server side is represented with two views represented as $\iota_+$ and $\iota_{++}$. The online network takes into account the former view while the target networks consider the latter view of the image. These views are obtained through the augmentation performed on the output of the decoder in the pre-training phase. These views undergo the encoder to extract the higher dimensional representation. Existing work such as MoCo-V2 (X. Chen et al., 2020) has validated the effectiveness of projection heads in the FL paradigm. The projection heads output the representations in lower dimensions. We used the same configuration of projection head as in existing studies, i.e., two linear layers with rectified linear unit (ReLU) activation. It is shown in Figure 2, that the local encoders share the weights with the online network only. Although the structure for both the target and online networks are the same the parameters of the target network need to be updated through exponential moving average via the online network. The formulation is shown in Equation 4.

$$\varrho_i = \theta . \varrho_{i-1} + (1 - \theta) . \varphi_i \quad (4)$$

The notation $\theta$ refers to the target decay which ranges within the bounds of [0,1]. The parameters of the online network and target network are represented by $\varphi$ and $\varrho$, respectively. Existing studies have suggested using larger values of $\theta$, i.e., around 0.9. The updates in the parameters of the online network are slow and steady till it reaches its optimal value. The similarity between the two views of the sample is calculated using InfoNCE (Oord et al., 2018) as shown in Equation 5.

$$Sim = -\log \frac{\exp\left(q_+ * \frac{q_{++}}{\mu}\right)}{\Sigma_0^N \exp\left(q_+ * \frac{q_-}{\mu}\right)} \quad (5)$$

where "*" refer to the cosine similarity between the views and $\mu$ represents the hyperparameter. A memory bank is used to store vectors of $N$ negative samples. The notation $q_+$ and $q_{++}$ are the predictions such that $q_+ = FC(Enc(\iota_+))$, where $FC$ is the fully connected layer. We consider $q_+$ and $q_{++}$ as a pair of positive examples and $q_-$ as a negative example. The images corresponding to negative examples are fed into both the encoder and the projection head to extract features followed by their storage in the memory bank. InfoNCE is then used to maintain the consistency between representations of the same images and the different ones. Based on the feature vectors a memory queue is generated that acts as the sequence of local negatives in a FIFO mechanism so that the contrast loss with each input sample is computed. The memory queue stores new features and recursively discards the older ones.

3.5 Aggregation Method and Local Network

At each communication round in phase 2, the online network on the server side transmits the encoder to the clients. At the client side in phase 2, a linear classifier is added, specifically a multilayer perceptron (MLP) is added for the classification task. The ground truth labels from the limited labeled dataset are used to compute the cross-entropy loss, which is widely used for computer vision tasks. For each client, the objective function is minimized, and weights are adjusted. Each client then sends the encoder part to the server for model aggregation. The server aggregates the encoder with the FC layers, which are then considered the updated online network. The updated network is then used for a subsequent round of training. The pre-training part in phase 1 mainly deals with data heterogeneity. However, to mitigate the label scarcity problem, we propose that the server should take into account the size of the dataset of the client and the frequency of client selection. The assumption is that the advantage of data associated with the client chosen several times dwindles, which limits the improvement concerning self-supervised learning. We propose an aggregation mechanism that maps the degradation relationship in order to improve the self-supervised learning process. Each client performs an update with respect to the formulation shown in Equation 6.

$$\varpi \leftarrow \varpi - \eta \nabla \mathbb{L}_{CE} \qquad (6)$$

where $\varpi$ represents the model in the client subset, currently, $\mathbb{L}_{CE}$ represents the cross-entropy loss, and $\eta$ refers to the learning rate. On the client side, individual gradients are computed to update the model. Considering that all clients send their updates to the server, the aggregation using the proposed method can be carried out as shown in equation 7.

$$\varpi_{i+1} \leftarrow \sum_{t=1}^{T} \frac{n_t}{n} \varpi_{t+1} \beta^{F_t} \qquad (7)$$

The term $F_t$ represents the frequency of a particular client that has participated in the aggregation process. The parameter $\beta$ introduces a gain-recession reaction suggesting that the higher the frequency of the client's participation, the lower the impact in the aggregation process. The notations $i$ and $t$ are used for iterative purposes, where as $T$ refers to the number of communication rounds.

4. **Experimental Setup and Results**

This section presents experimental analysis and results on IoMT, specifically medical image datasets to validate the efficacy of the SelfFed framework. First, the details regarding the datasets are laid out followed by the experimental setup for the SelfFed framework. We present an analysis to show the robustness of the proposed work for data heterogeneity and label scarcity issues and provide a comparative analysis with existing works, accordingly.

4.1 Datasets

We validate the performance of the SelfFed method on two medical image-related tasks. The first is the classification of COVID-19 and pneumonia from chest X-rays, and the second is to use retinal fundus images for diabetic retinopathy detection. Both of the tasks differ in terms of label distributions, image acquisition, and image modalities, respectively. For instance, chest X-rays are acquired using X-ray scanners while the fundus images are captured using specialized cameras.

4.1.1 COVID-19 Dataset

For the COVID-19 dataset, we adopt COVID-FL (Yan et al., 2023) which represents federated data partitions in real-world settings. The COVID-FL comprises over 20,000 chest X-ray images curated from eight data repositories that are available in the public domain. These 8 repositories include the Cohen Dataset[1] (we removed the duplicate images), the Guangzhou pediatric dataset (Kermany et al., 2018), the RSNA pneumonia detection challenge dataset[2], the RSNA COVID-19 open radiology database (International)[3], the European Society of Radiology dataset[4], the Italian Society of Medical and Interventional Radiology COVID-19 database[5], German institute for Diagnostic and Interventional Radiology dataset[6], and the Valencia Region Image bank database (Kermany et al., 2018). Each of the eight repositories is represented as a data site to mimic a medical institution in the FL paradigm. It also complies with data heterogeneity and label scarcity issues as classes might vary in the dataset corresponding to specific sites. Furthermore, the acquisition machines that obtained X-ray images were different, along with the patient populations. Therefore, the dataset naturally emulates the data heterogeneity and label scarcity issues. We follow the same protocol as in (Yan et al., 2023), such that the split ratio for training and testing is set to 80% and 20%, which yields around 16,000 training and 4,000 testing images. The proportion of the data at each site is homogeneous. Each client (i.e., hospital) is considered to be a test set that can be held out for evaluation.

4.1.2 Retina Dataset

We also undertake the Diabetic Retinopathy Competition dataset[7] from Kaggle. The dataset comprises over 35,000 fundus images obtained using multiple specialized cameras. The original dataset divides the images into 5 categories, i.e., proliferating, severe, moderate, mild, and normal. However, in this study we have binarized the labels, resulting in 2 categories: diseased and normal. 9,000 balance images are randomly selected from the dataset for the training set while 3,000 images were selected for the testing set, accordingly.

4.2 Experimental Setup
4.2.1 Preparation of Non-IID Dataset

For the COVID-FL dataset, they already manifest skewness in feature and label distribution as in real-world settings. It has been observed in the COVID-FL dataset that some clients contain large volumes of data, which is significantly higher than the other clients. In this regard, the clients having skewed label distribution are partitioned into sub-clients while making sure that the instances do not overlap among the sub-clients. After such partitioning, the total number of clients stands at 12 for the COVID-FL dataset.

---

[1] https://github.com/ieee8023/covid-chestxray-dataset
[2] https://www.kaggle.com/c/rsna-pneumonia-detection-challenge/
[3] https://doi.org/10.7937/91ah-v663
[4] https://eurorad.org/
[5] https://www.sirm.org/category/senza-categoria/COVID-19/
[6] https://github.com/ml-workgroup/covid-19-image-repository
[7] https://www.kaggle.com/c/diabetic-retinopathy-detection

For the Retina dataset, we consider Dirichlet distribution as suggested in (Yan et al., 2023) to model non-IID and IID characteristics. It has also been suggested that simulated partitions in the dataset provide greater freedom for investigation as it's easy to manipulate them and varying degrees of testing can be performed concerning heterogeneity of the data. A dataset with $\Omega$ classes and $M$ local clients can be randomly partitioned through simulation as shown in equation 8.

$$\rho_i = \{\rho_{i,1}, \ldots, \rho_{i,M}\} \sim Dir_M(\delta) \tag{8}$$

where the $L_1$ norm of $\|\rho_i\| = 1$. A proportion of $i^{th}$ class instances are assigned to $N^{th}$ client. The degree of heterogeneity is controlled by the parameter $\delta$ in equation 8, suggesting that higher values of $\delta$ lead to a lower degree of heterogeneity. We consider three values of $\delta$ that simulate the level of data heterogeneity from IID to moderate non-IID and severe non-IID, respectively. The values of $\delta$ are selected to be 100, 1.0, and 0.5, for the Retina dataset with 5 number of clients, accordingly. The dataset was split with $\delta = 100$ is referred to as IID (Split1), while the split with values 1.0 and 0.5, is referred to as moderate non-IID (Split2) and severe non-IID (Split3).

### 4.2.2 Data Augmentation

The data augmentation parameters vary slightly in the pre-training and fine-tuning phases. Random horizontal flipping and crop patches of size $224 \times 224$ along with random scaling are performed in both phases. The random color jittering is performed during pre-training while random rotation (10 degrees) is performed during fine-tuning. For Retina and COVID-FL datasets in the pre-training phase, the random scaling factor is selected from the values [0.2, 1.0] and [0.4, 1.0], respectively. Similarly, the random scaling factor for both the dataset in fine-tuning phase is selected from the values [0.8, 1.2] and [0.6, 1.0], respectively.

### 4.2.3 Pre-training and Fine-tuning Setup in SelfFed

The simulations for this study have been carried out using DistributedDataParallel (DDP) module and PyTorch framework. The backbone for the proposed method is chosen to be Swin Transformer (Liu et al., 2021). For the MAE, 16x16 patches were selected. Augmentative masking was performed for 60% of the image patches in a random manner. AdamW is used as an optimizer in the proposed method with its default values. The same hyperparameters are used for both federated and centralized learning. The batch size and learning rate were varied several times to ensure hyperparameter tuning, the values that achieve the best results are reported below. For the Retina dataset, the batch size and learning rate were set to 128 and $1e^3$, while for the COVID-FL dataset, the parameters were set to 64 and $3.75e^{-4}$. The cosine decay for SelfFed-MAE was set to 0.05. The SelfFed-MAE uses a warmup period of five epochs to run along with 1.6k communication rounds. We train the model with a batch size of 256 and a learning rate of $3e^{-3}$ for the Retina dataset, and a batch size of 64 with the same learning rate for the COVID-FL dataset. The fine-tuning is performed for 100 communication rounds, respectively. The hyperparameter value for $\beta$ was set to be 0.95. We validate our approach using accuracy as an evaluation metric. The choice of evaluation metric is in compliance with existing studies.

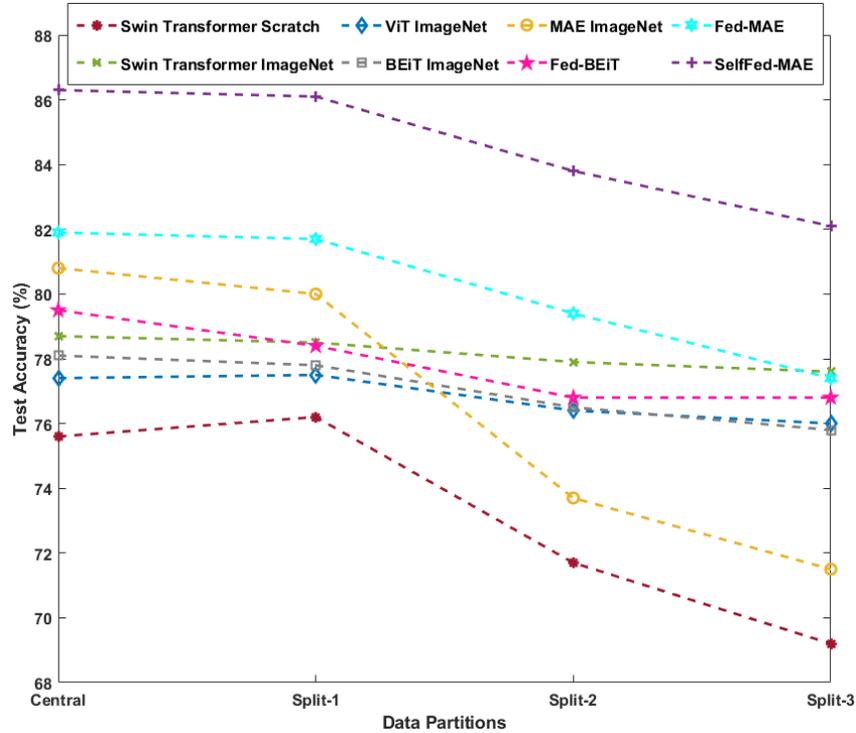

Figure 6 Comparative analysis with respect to accuracy on Retina Dataset to show the efficacy of SelfFed for data heterogeneity.

## 4.3 Results for Data Heterogeneity

To validate the efficacy of the SelfFed-MAE approach, we compare our results with some baseline methods that include Swin Transformer (from scratch), ImageNet pre-training with Swin Transformer, ImageNet pre-training with ViT, ImageNet pre-training with BEiT, and ImageNet pre-training with MAE, and an existing works Fed-MAE and Fed-BEiT (Yan et al., 2023) . For both datasets, the baseline approaches are pre-trained on ImageNet22K (Deng et al., 2009) in a centralized manner, while the Swin Transformer (from scratch) is trained in a decentralized manner. Additionally, for COVID-FL, we compare our approach with pre-trained Swin Transformer, ViT, BEiT, and MAE, on the CXR14 dataset (X. Wang et al., 2017) which comprises around 112K images. We run the fine-tuning process for 1K communication rounds and 100 communication rounds for the networks trained from scratch and for pre-trained networks, respectively. The results are reported in Figures 6, 7, and 8. The results show that the SelfFed-MAE is consistent in terms of results on both datasets concerning the data heterogeneity problem. Methods such as MAE and BEiT pre-trained on ImageNet show a large discrepancy (drop) in accuracy. Even when the heterogeneity level of the data is severe the proposed method yields not only the best accuracy but also a consistent hold over the performance degradation. We observe maximum accuracy improvement on Retina and COVID-FL datasets, i.e., 8.8% and 4.1%, using SelfFed-MAE over pre-trained networks, respectively. The reason that the proposed method outperforms the existing works, especially the supervised learning strategy is that the domain shift between the ImageNet and the Retina or COVID-FL dataset is quite large, therefore the fine-tuning does not help much to the networks pre-trained on ImageNet. The works Fed-MAE and Fed-BEiT (Yan et al., 2023), show that they perform well in terms of distributed settings as the domain shift for this method decreases because of the contrastive learning strategy, however, the use of Swin transformer as an encoder and the novel aggregation strategy sets the proposed method apart in terms of performance. The performance provides a firm basis for the proposed work to be fit for real-world medical applications as it performs better even with the much smaller size of the dataset in comparison to other centralized learning settings. Furthermore, the real-world applicability of the proposed method is also highlighted by its performance on non-IID settings, which clearly represent different data distributions across different hospitals.

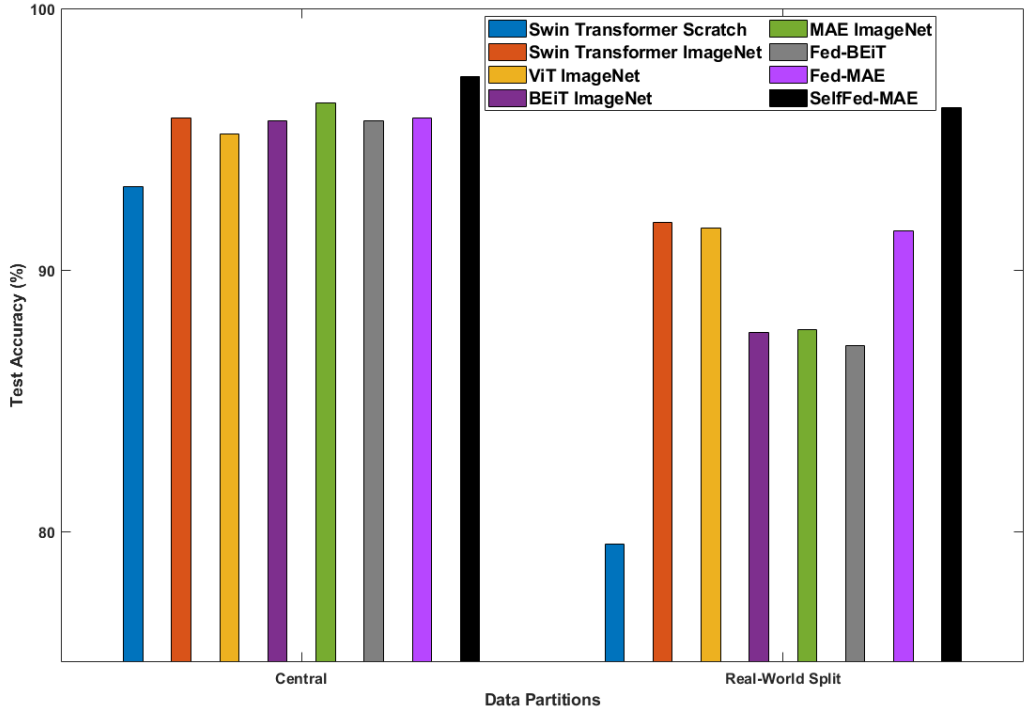

Figure 7 Comparative analysis with models pre-trained on ImageNet using accuracy on COVID-FL Dataset to show the efficacy of SelfFed for data heterogeneity.

4.4 Results for Label Scarcity

In order to prove that the SelfFed framework can provide improved performance even with a limited number of labels, we conducted the following experiment on Retina Dataset. We use 10%, 30%, and 70% of labeled samples at the SelfFed fine-tuning stage, which results in 1,000, 3,000, and 6,000 labeled images, respectively. The results of this experiment are reported in Figure 9. Even with 10% of the labeled samples, our method achieves better performance in comparison to other baselines while using 100% of the labeled data in both IID and non-IID settings. The results for training Swin Transformer from scratch are compliant with the existing works as it does not perform well under a smaller number of labeled samples.

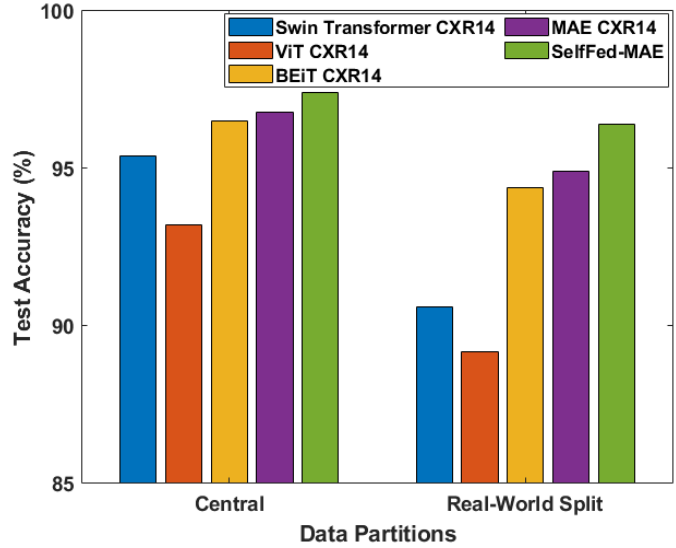

Figure 8 Comparative analysis with models pre-trained on CXR14 using accuracy on COVID-FL Dataset to show the efficacy of SelfFed for data heterogeneity.

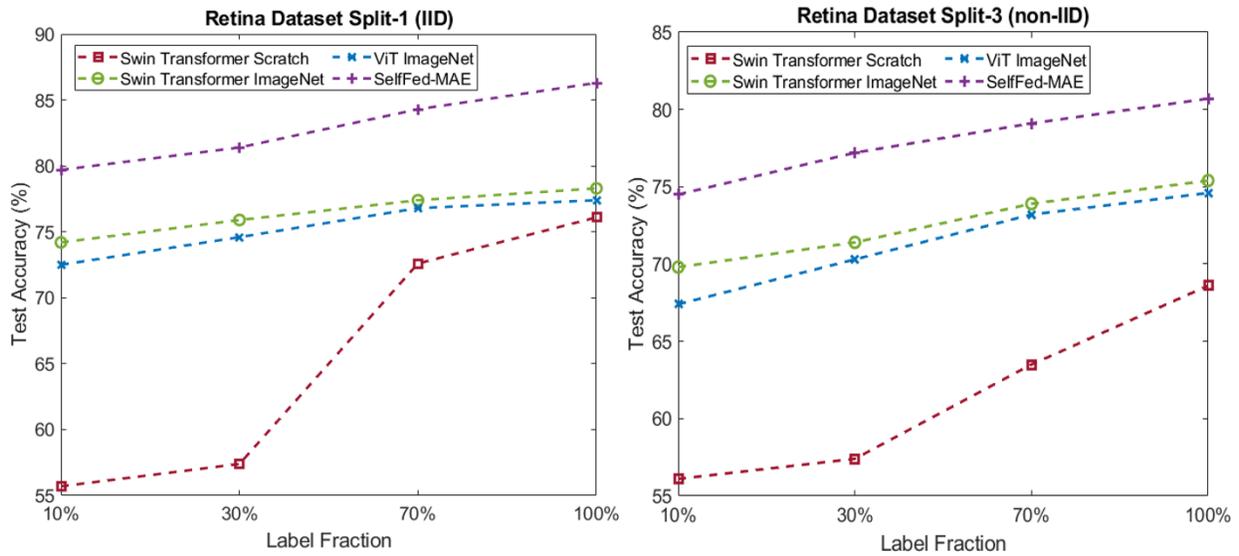

Figure 9 Efficacy of the SelfFed-MAE concerning varying fractions of labeled samples under IID and non-IID settings.

### 4.5 Results for Ablation Study

In this subsection, we provide the results with respect to different components of the proposed architecture to show that each of the proposed components contribute to the performance gain. The ablation study is carried out on Retina dataset using only Split-3. We first perform the ablation study on the data augmentation methods using Test Accuracy, Centralized learning setup with ViT backbone (to show the performance comparison between centralized and contrastive learning), Encoder (to show the performance comparison between ViT and Swin Transformer) and Aggregation methods, using the test accuracy, respectively. The results are reported in Table 2. The results clearly illustrate that the data augmentation method, contrastive learning strategy, Swin transformer encoder, and proposed aggregation mechanism contribute to the overall performance gain of the system.

Table 2 Ablation study for the performance analysis of Data Augmentation, Contrastive Learning, Employed encoder, and Aggregation methods, respectively.

| Data Augmentation | Test Accuracy |
|---|---|
| w/o augmentation | 76.4 |
| Horizontal flipping + random cropping | 81.5 |
| Color Jitter + Scaling + Horizontal flipping + random cropping | 82.1 |
| **Learning Strategy** | **Test Accuracy** |
| Centralized Learning (w ViT backbone) | 76.0 |
| Centralized Learning (w Swin Transformer backbone) | 77.6 |
| SelfFed-MAE (w ViT backbone) | 79.4 |
| SelfFed-MAE (w Swin Transformer backbone) | 82.1 |
| **Aggregation Mechanism** | **Test Accuracy** |
| FedMoCov3 (X. Chen et al., 2021) | 77.1 |
| FedMoCo (K. He et al., 2020) | 74.8 |
| FedBYOL (Grill et al., 2020) | 77.6 |
| FedAvg (McMahan et al., 2017) | 74.6 |
| Fed-BEiT (Yan et al., 2023) | 76.8 |
| Fed-MAE (Yan et al., 2023) | 77.4 |
| Proposed | 82.1 |

4.6 Results for sensitivity for the parameter $\beta$

As we propose a novel aggregation strategy for self-supervised federated learning paradigm, specifically for medical images, it is important to provide the sensitivity analysis for the hyperparameter $\beta$ (the weighting adjustment factor). We show the sensitivity analysis results for the parameter $\beta$ in Figure 10. It can be visualized that the selection of weight adjustment factor has a significant impact on the accuracy of the proposed method. To show the impact of parameter $\beta$ on the accuracy in both IID and non-IID settings, we vary the values of $\beta$ as 0.6, 0.75, 0.9, 0.95, 0.99, 0.999, and 1. We report the results on the split-3 of the Retina dataset. If we chose the value of $\beta$ as 1, the intrinsic characteristics of the aggregation method becomes similar with FedAvg. Our method shows that selecting the value below 0.9 significantly decreases the test accuracy, which would eventually lead to the addition of more clients. Our methods achieve the best results for client aggregation when the $\beta$ value is selected to be 0.95.

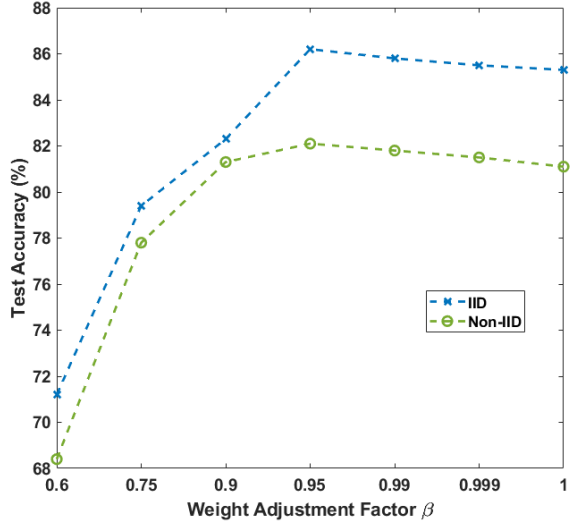

Figure 10 Sensitivity analysis for the parameter $\beta$

## 5. Limitations

One of the downsides to using Swin Transformers as encoders in the proposed study is the number of parameters for SelfFed-MAE, which is around 130.7 M parameters. In comparison, the contrastive learning ViT when used in conjunction with MAE yields around 112 M parameters. The size of the model plays an impact when transmitted for the fine-tuning process as the model with larger parameters tends to add delay. However, in the defense of the proposed method, the communication cost would not be affected much as the proposed method achieves a better performance in less communication rounds when compared to the other methods. We believe it is important to highlight the limitations of the proposed work, however, we assume that the limitation does not affect the real-world applicability of the proposed work.

## 6. Conclusion

This paper addresses an understudied scenario in the FL paradigm where both the data heterogeneity and label scarcity issues are handled simultaneously. We proposed a SelfFed framework that used augmentative modeling by leveraging the MAE method to deal with data heterogeneity. We also propose a novel contrastive network that handles two views of an instance and propose a novel aggregation strategy that reduces the impact of a client who participated various times in the fine-tuning process. The fine-tuning stage comprising our contrastive network and novel aggregation strategy overcomes the label scarcity issue. We illustrate with our experimental results that the proposed SelfFed is efficient when using non-IID data and performs better than existing baselines such as Swin Transformer pre-trained on ImageNet, ViT pre-trained on ImageNet, MAE pre-trained on ImageNet, and BEiT pre-trained on ImageNet. Our results also indicate that the SelfFed framework is successful in handling label scarcity while outperforming existing baselines to the best of our knowledge.

The proposed SelfFed framework shows promising results in the domain of medical imaging in terms of performance. As a future direction, we would like to test the SelfFed framework on MRI and CT Scan images to explore the domain adaptability and scalability of the proposed approach. We also intend to explore the security issues concerning the proposed method, specifically the ones related to model security attacks, such as model inversion, gradient leakage, model poisoning, and membership inference attacks, which can affect the applicability of the proposed work (Khowaja et al., 2022, 2024).